\documentclass[authorversion, dvipsnames,format=sigconf]{acmart}

\AtBeginDocument{%
  }

\setcopyright{acmlicensed}

\copyrightyear{2026}
\acmYear{2026}
\setcopyright{cc}
\setcctype{by}
\acmConference[GECCO Companion '26]{Genetic and Evolutionary Computation Conference}{July 13--17, 2026}{San Jose, Costa Rica}
\acmBooktitle{Genetic and Evolutionary Computation Conference (GECCO Companion '26), July 13--17, 2026, San Jose, Costa Rica}
\acmDOI{10.1145/3795101.3805286}

\usepackage{siunitx}
\usepackage{lipsum}
\usepackage{hyphenat}
\usepackage{graphicx}
\usepackage{subcaption}
\usepackage{bm}
\usepackage{multirow}
\usepackage{amsmath}
\usepackage{comment}
\usepackage{makecell}

\definecolor{dark-red}{rgb}{0.4,0.15,0.15}
\definecolor{dark-blue}{rgb}{0.15,0.15,0.8}
\definecolor{medium-blue}{rgb}{0,0,0.5}
\hypersetup{
    colorlinks, linkcolor={dark-red},
    citecolor={dark-blue}, urlcolor={medium-blue}
}

\begin{document}

\title[The Evaluation Cost of Task Specialization in Evolutionary Multi-Robot Systems]{The Evaluation Cost of Task Specialization in \\ Evolutionary Multi-Robot Systems}

\author{Paolo Leopardi}
\email{[firstname].[lastname]@uni-konstanz.de}
\orcid{0009-0005-7064-9344}
\author{Heiko Hamann}
\orcid{0000-0002-2458-8289}
\author{Jonas Kuckling}
\orcid{0000-0003-2391-2275}
\affiliation{%
  \institution{Centre for the Advanced Study of Collective Behaviour \&\\
  Department of Computer and Information Science\\
  University of Konstanz}
  \city{Konstanz}
  \country{Germany}
}

\author{Tanja Katharina Kaiser}
\email{tanja.kaiser@utn.de}
\orcid{0000-0002-1700-5508}
\affiliation{%
 \institution{
 Department of Computer Science and Artificial Intelligence\\
 University of Technology Nuremberg}
 \city{Nuremberg}
 \country{Germany}}

\renewcommand{\shortauthors}{Leopardi et al.}

\begin{abstract}
Task specialization can improve the efficiency of multi-robot systems (MRSs). 
Previous works have investigated the emergence of task-specialist robot controllers through evolutionary optimization and have argued that task specialization is more likely to evolve when subtask behaviors are readily available as building blocks. 
However, the available evaluation budget must be distributed across all subtasks, whereas a single generalist behavior can exploit the entire budget for its own optimization.
We present a cost-benefit analysis of evolving task-specialist versus generalist behaviors in a foraging scenario here. 
In a physics-based robotics simulator, we study the total evaluation budget required to evolve task-specialist behaviors that outperform generalist behaviors across MRS sizes. 
We show that with increasing MRS size, a lower total evaluation budget is sufficient to evolve specialists that outperform generalists.
\end{abstract}

\ccsdesc[300]{Computing methodologies~Multi-agent systems}
\ccsdesc[500]{Computer systems organization~Evolutionary robotics}

\keywords{Evolutionary Robotics, Evolutionary Multi-Robot Systems, Task Specialization, Division of Labor}

\maketitle

\section{Introduction}

\begin{figure}[t]
    \centering
    \includegraphics[width=\linewidth]{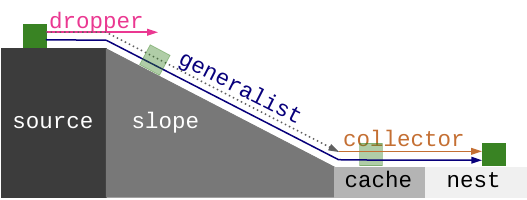}
    \caption{Task overview: objects (green squares) have to be transported from the source to the target. Robots can either execute the full task (generalist) or one of two subtasks: pushing objects down the slope into the cache (dropper) and transporting objects from the cache to the target (collector).}
    \label{fig:task_overview}
    \Description{} 
\end{figure}

Division of labor is characterized by task specialization among subgroups within a multi-robot system (MRS), each responsible for a distinct task. 
Distributing tasks can increase overall system efficiency~\cite{lerman2002mathematical} and
simplify individual robot tasks.
However, when subtasks are interdependent, coordination costs can negatively affect system performance~\cite{brutschy2014self} and weaken synergies in robot-environment interactions~\cite{nolfi00}.

Subtask behaviors are often hand-coded, but can also be optimized using learning-based approaches~\cite{albrecht2024multi,bongard13,trianni08}.
In the latter case, task specialization can also emerge without being explicitly enforced~\cite{vanDigg2024emergence,ferrante2015}. 
When subtask behaviors are optimized independently, each one requires its own evaluation budget. 
Given a fixed evaluation budget~\(E\) for the overall task, task specialization entails evolving \(n\)~specialized controllers for the \(n\)~subtasks, reducing the available budget per controller, for example, to~\(E/n\).  

We investigate under which conditions evolving task\hyp{}specialist behaviors yield superior team performance compared to evolving generalist behaviors within a fixed total evaluation budget~\(E\)~\cite{Doncieux2015,trianni08}.
We assume that the overall task has been manually decomposed into specialized, interdependent subtasks, that can be performed in separate areas, and a predefined allocation of tasks to robots. 
Our focus is thus on the evolution of behaviors for the pre-partitioned subtasks, rather than on task partitioning or allocation. 

Our benchmark is an object retrieval task inspired by the foraging of leafcutter ants~\cite{ferrante2015,hart2002task}.
Objects must be transported from a source across a slope and a cache area to the target area. 
System efficiency can be increased by partitioning the task into two subtasks and leveraging the slope. 
However, subtasks are sequentially interdependent as collecting robots can only retrieve an object to the target when another robot has deposited it in the cache beforehand. 

In this paper, we study the total evaluation budget~\(E\) required for optimizing task-specialist behaviors that outperform generalist behaviors across varying MRS sizes.
We show that larger MRS sizes lead to smaller required total evaluation budgets~\(E\).

\section{Related Work}

Division of labor in MRSs is widely studied, with object retrieval scenarios such as foraging commonly used as benchmarks~\cite{ferrante2015,montanier2016behavioral,Pini2014}.
Factors such as spatial separation, MRS size, and resource distribution have been identified to contribute to the emergence of task specialization~\cite{montanier2016behavioral}.
We built on these findings by testing various MRS sizes in our experiments.

The ``economic-transport hypothesis''~\cite{roschard2003cutters} posits that task specialization is promoted when environmental features can be exploited to improve overall task efficiency. 
This occurs, for example, when object transport can be sped up by exploiting a slope in the environment~\cite{ferrante2015}, as in our experiments. 
Ferrante \textit{et al.}~\cite{ferrante2015} evolved division of labor in a homogeneous robot swarm \textit{de novo} by rewarding group performance using grammatical evolution and predefined low-level behavioral primitives.  
The authors argue that task specialization is more likely to emerge when behaviors for subtasks have already been evolved or are readily available as behavioral building blocks. 
Our work focuses on the evolutionary optimization of the unpartitioned task and the pre-partioned subtaks, which could serve as such pre-adapted behavioral building blocks.

To the best of our knowledge, we are the first to study the optimization costs of generalist versus task-specialist behaviors.
Our core question is not which factors promote the emergence of task specialization itself, but rather, what evaluation budget is required to optimize pre-partitioned task-specialist behaviors that outperform generalist behaviors in different settings.

\section{Method}

\subsection{Task Description}
\label{sec:task_description}

We study a foraging task in which \(S\)~robots have to retrieve objects to a target area. 
The environment has four areas~\(\mathcal{A}\): source, slope, cache, and target (see Fig.~\ref{fig:task_overview}). 
In the arena are \(M\)~retrievable objects~\(O_m, \, m \in \{1, \dots, M\}\), which are initially distributed in the source.
Objects disappear when they are deposited in the target, and a new object appears in the source area.
When on the slope, objects will automatically slide into the cache. 
The performance of the MRS is measured by the number of objects~\(C^T_\text{target}\) transported to the target within a given time period~\(T\)~\cite{balch2002taxonomies}, as given by 
\begin{align} \label{equ:c_T}
    C^T_\text{target} &= \sum_{t=1}^{T} \sum_{m=1}^M H(O_m, t, \text{target}), \\
    H(O_m, t, \mathcal{A}_i) &= 
    \begin{cases}
    1,& \text{if object \(O_m\) is in area \(\mathcal{A}_i\) at time~$t$} \\
    0,              & \text{otherwise} 
\end{cases}, 
\end{align}
with \(H(O_m, t, \mathcal{A}_i)\) indicating whether object~\(O_m\) was located in area~\(\mathcal{A}_i\) at time~\(t\).
The task can be carried out using two distinct strategies: (i)~in the \textit{generalist} approach, each robot independently executes the entire object retrieval task, and (ii)~in the \textit{specialist} approach, robots divide the task into two interdependent subtasks.  
\textit{Droppers} transfer objects from the source to the cache by exploiting the slope and \textit{collectors} transport the objects from the cache to the target.
The number of subtasks~\(n\) for each strategy is \(n=1\) for generalists and \(n=2\) for specialists. 
For the specialist strategy, we enforce a subtask-balanced fifty-fifty allocation of MRS size~\(S\) to the pre-partitioned \(n=2\)~specialist behaviors, resulting in subteam size \(S^*={S}/{n}\). 
We choose a balanced allocation as both subtasks operate over areas of equal size.

\subsection{Experimental Setup}

We run our experiments in an arena of size~\qty{2}{\meter} $\times$ \qty{6}{\meter} in the open-source simulator ARGoS~\cite{pinciroli2012argos}. 
Target, cache, and source are \qty{1}{\meter} long, and the slope is \qty{3}{\meter} long.
Three light sources are located above the target.
We use \(M=10\) cylinders (radius:~\qty{0.05}{m}, height:~\qty{0.08}{m}, weight:~\qty{0.1}{\kilo\gram}) as objects. 
We employ teams of differential-drive foot-bot robots~\cite{bonani2010marxbot} that are equipped with 
\(24\)~light sensors, \(24\)~proximity sensors, and \(4\)~ground sensors each.
We extend the platform with a 2D~LiDAR providing~\(360^\circ\) coverage, which is divided into eight sectors represented by their minimum range value.
Objects are detectable only by the proximity sensors, while robots are detectable by the proximity sensors and the LiDAR.
We normalize all sensor values to~\([0,1]\).
Robots can only push objects, and consequently, they may lose physical contact when an object slides down the slope.
However, they can easily relocate the object in the cache by driving straight downhill.
The robot's speed ranges from \(0\) to \(1~\unit{\meter/\second}\).
On the slope, the speed is scaled by~\(0.2\) when heading uphill and by~\(1.5\) when heading downhill.

Each robot is controlled by a fully-connected feedforward artificial neural network (ANN) with one hidden layer of eight neurons.
Controllers share the same weights.
The ANN outputs the velocities of the right~\(v_r\) and left~\(v_l\) wheels.
The ANN inputs are the \(24\)~light sensor readings, \(24\)~proximity sensor readings, \(4\)~ground sensor readings, and the \(8\)~LiDAR values.

\subsection{Evolution} \label{sec:evolution}

\begin{table}
    \centering
    \caption{Behavior-specific evolutionary setup parameters} 
    \begin{tabular}{ccccc}
        \hline
        \textbf{Behavior} & \(\mathbf{E^*}\) & \textbf{\makecell{Initial\\Object Area}} & \textbf{\makecell{Goal\\Object Area}} & \(\mathbf{S^*}\) \\ \hline
        
       Generalist & 5000 & Source & Target & \(\{2,4,6,8\}\) \\ \hline
        Dropper   & 2500 & Source & Cache  & \(\{1,2,3,4\}\) \\ 
        Collector  & 2500 & Cache & Target  & \(\{1,2,3,4\}\)\\ \hline
    \end{tabular}
    \label{tab:evolutionary_setup_parameters}
\end{table}

We optimize the ANN weights using an evolutionary algorithm~\cite{eiben2015introduction}.
A genome directly represents the ANN weights as floats that are initially uniformly sampled from~\([-1,1]\).
Our algorithm uses tournament selection with tournament size~\(5\), no elitism, one-point crossover with \SI{40}{\%}~probability, and a population size of~\(96\).
Each gene is mutated with \SI{20}{\%}~probability, using Gaussian mutation with zero mean and standard deviation of~\(0.2\).
The evaluation time~\(T_\text{eval}\) is \(60\)~seconds and the evaluation budget~\(E\) is \(5000\) (i.e., the number of generations for evolving a strategy). 
The three different behaviors for the two foraging strategies are evolved independently.
The effective evaluation budget for each behavior is computed as \(E^*={E}/{n}\).
For the generalist strategy, \(E^* = E\) as the entire MRS performs only one behavior.
For the specialist strategy, \(E^* = 5000/2 = 2500\) as we evolve dropper and collector behaviors.
The robot's initial poses are uniformly randomized across the entire arena.
The initial object positions and the goal area to which an object must be delivered depend on the behavior being evolved, see Tab.~\ref{tab:evolutionary_setup_parameters}.

Task-level fitness for the generalist~\(F_G\) and the collector~\(F_C\) behaviors is given by
\begin{equation} 
    F_G = F_C = C^T_\text{target} = \sum_{t=1}^{T_\text{eval}} \sum_{m=1}^M H(O_m, t, \text{target}).
    \label{eq:F_G}
\end{equation}
For the dropper behavior, we reward the number of objects pushed into the cache by the MRS, as given by
\begin{equation}
    F_D = \sum_{t=1}^{T_\text{eval}} \sum_{m=1}^M H(O_m, t, \text{cache}).
    \label{eq:F_D}
\end{equation}
Eqs.~\ref{eq:F_G} and~\ref{eq:F_D} are aggregate fitness functions~\cite{nelson2009fitness} since they measure only task completion, regardless of the behavior exhibited.
We evolve controllers for MRS sizes~\(S \in \{2,4,6,8\}\) and run five evolutionary independent runs per behavior and subteam size~\(S^*\).

\subsection{Post-Evaluation} 
\label{sec:evaluation_post-evaluation}

We post-evaluate the performance of the evolved robot controllers for the two strategies by comparing \(C^T_\text{target}\) (Eq.~\ref{equ:c_T}) for post-evaluation time~\(T_\text{p-eval}\) of~\SI{60}{s} and test for statistical significance using the Mann-Whitney U~test~\cite{mann1947test}, with \(p\)-values adjusted via the Holm–Bonferroni correction~\cite{holm1979simple}.
We compare the performance of equal MRS sizes~\(S\) by combining droppers and collectors evolved with subteam size~\(S^* = S/2\) for the specialist strategy.
Task-specialist MRSs are put together by combining the best evolved dropper with the best evolved collector in the order of the evolutionary runs.
 
We evaluate the best evolved individuals of different evaluation budgets~\(\tilde{E} \in \{10, 20, \ldots, 90, 100, 200, \ldots, 900, 1000, 1500, \ldots, 5000\}\). 
For each evaluation budget~\(\tilde{E}\), we evaluate the controller defined by the best evolved genome of the generation~\(\tilde{E}/{n}\) in 20~random trials.
We define the break-even point as the evaluation budget at which the median performance of the specialists equals or exceeds that of the generalists and remains higher or equal thereafter.

\section{Results}

\subsection{Evolution}
\label{sec:post-evaluation_fitness}

\begin{figure*}
    \centering
    \begin{subfigure}[t]{0.32\linewidth}
        \centering
        \includegraphics[width=\linewidth]{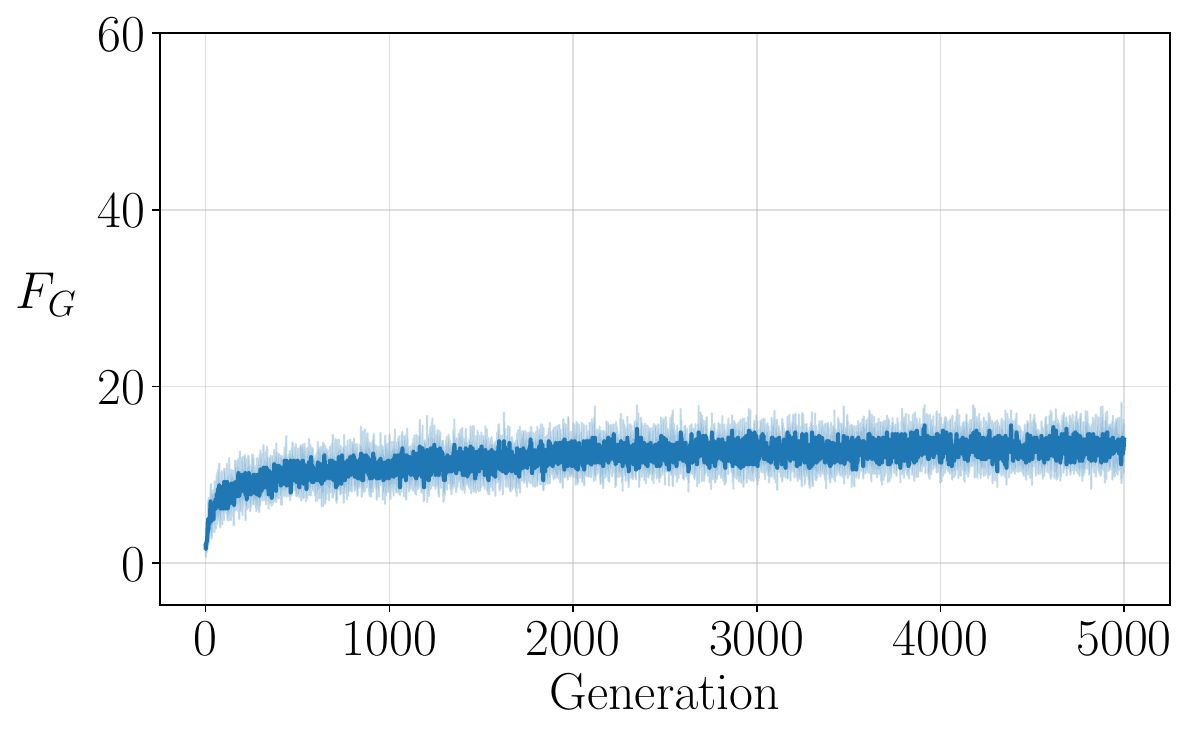}
        \caption{generalist}
        \label{subfig:generalist_10}
    \end{subfigure}    
    \hfill
    \begin{subfigure}[t]{0.32\linewidth}
        \centering
        \includegraphics[width=\linewidth]{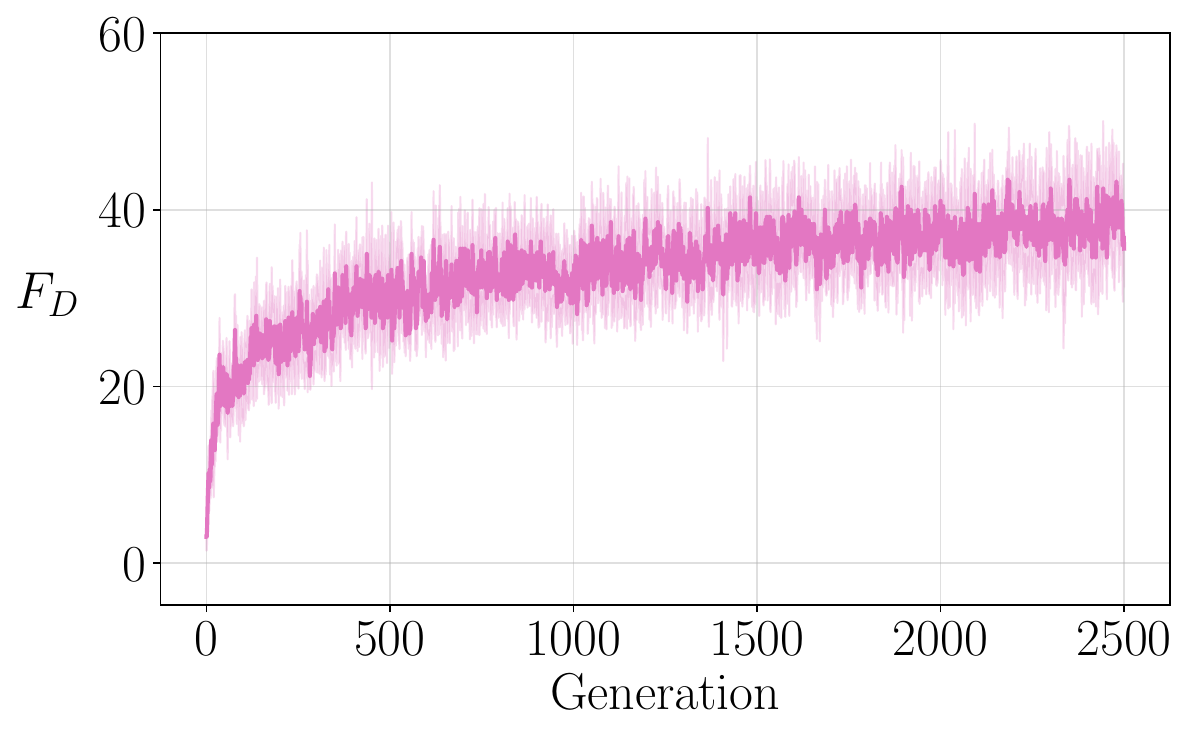}
        \caption{dropper}
        \label{subfig:dropper_10}
    \end{subfigure}
    \hfill
    \begin{subfigure}[t]{0.32\linewidth}
        \centering
        \includegraphics[width=\linewidth]{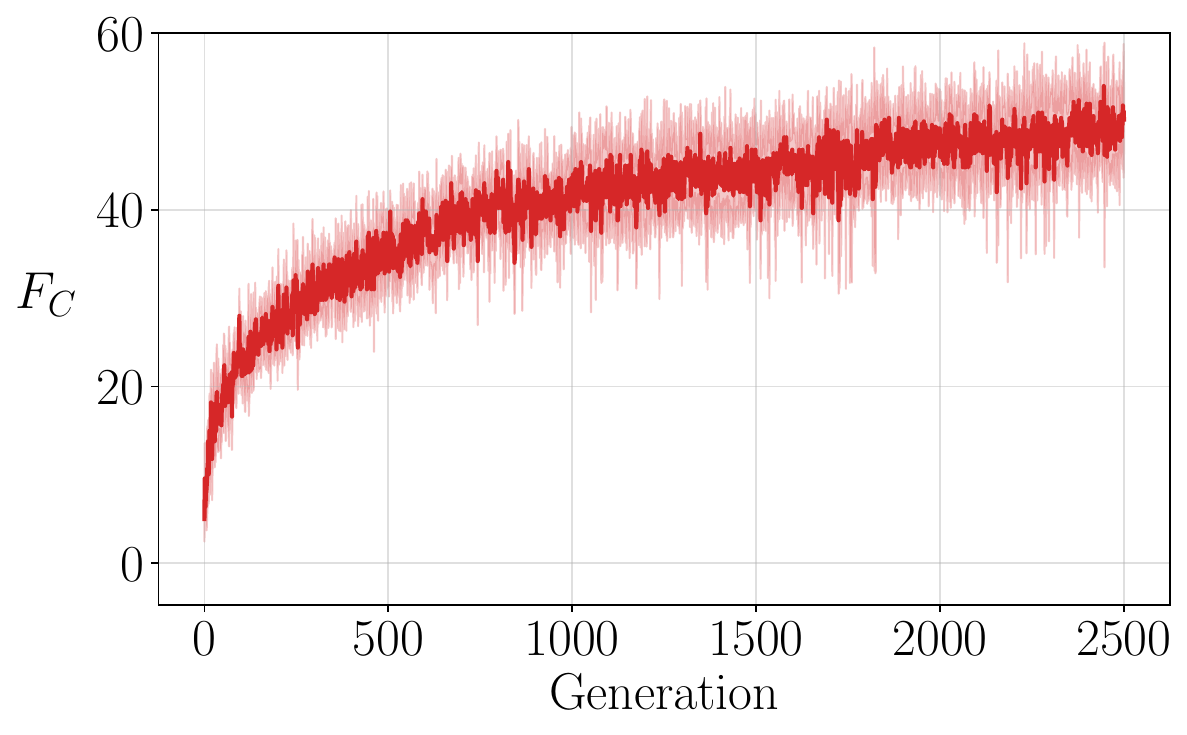}
        \caption{collector}
        \label{subfig:collector_10}
    \end{subfigure}
    \hfill
    \caption{Best fitness~\(\mathbf{F_G}\), \(\mathbf{F_D}\) and \(\mathbf{F_C}\) (Eqs.~\ref{eq:F_G}, \ref{eq:F_D}) for generalist (\(\mathbf{S^* = S =4}\)), dropper (\(\mathbf{S^*=2}\)), and collector (\(\mathbf{S^*=2}\)) behaviors with \(\mathbf{M = 10}\)~objects over 5 independent runs (solid line: mean, shaded area: standard deviation).} 
    \label{fig:evolution_fitness_5_objects}
    \Description{}
\end{figure*}
\begin{figure*}
    \centering
    \begin{subfigure}[t]{0.24\linewidth}
        \centering
        \includegraphics[width=\linewidth]{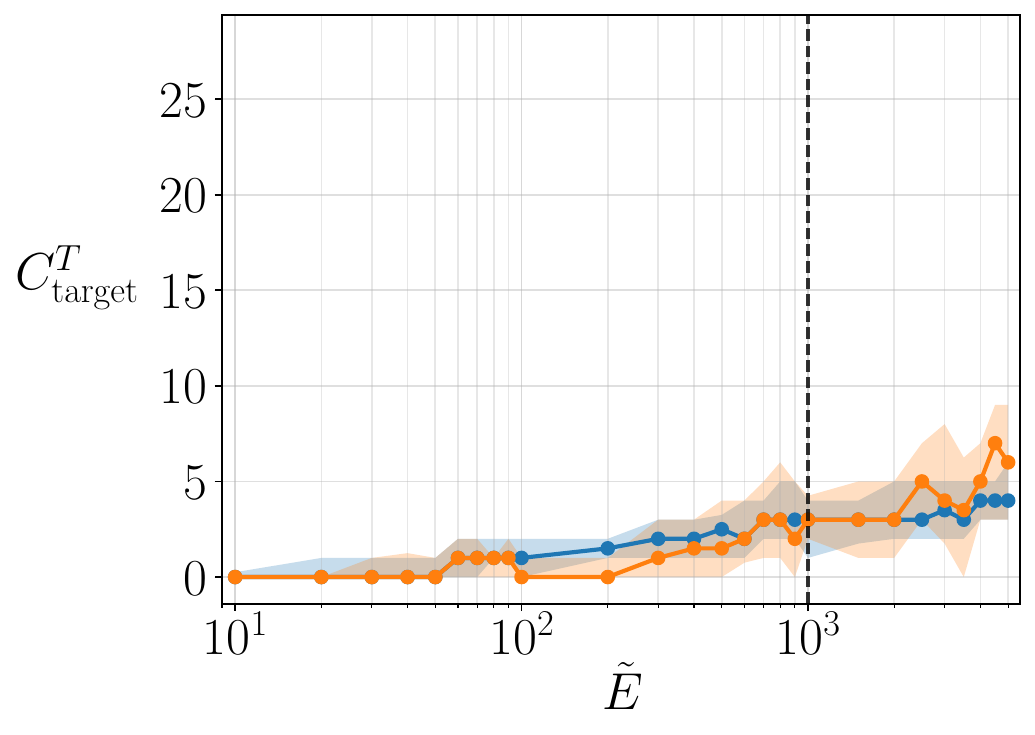}
        \caption{\(S=2\)}
        \label{fig:swarm_size_2}
    \end{subfigure}
    \hfill
    \begin{subfigure}[t]{0.24\linewidth}
        \centering
        \includegraphics[width=\linewidth]{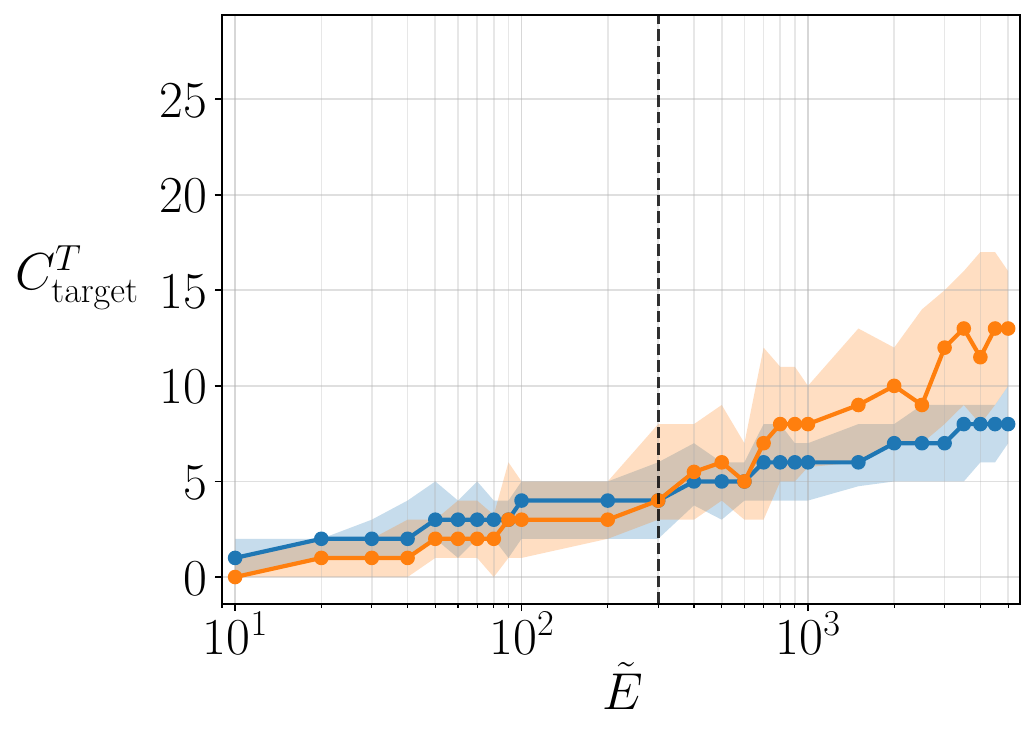}
        \caption{\(S=4\)}
        \label{fig:evaluation_swarm_size_S4}
    \end{subfigure}
    \hfill
    \begin{subfigure}[t]{0.24\linewidth}
        \centering
        \includegraphics[width=\linewidth]{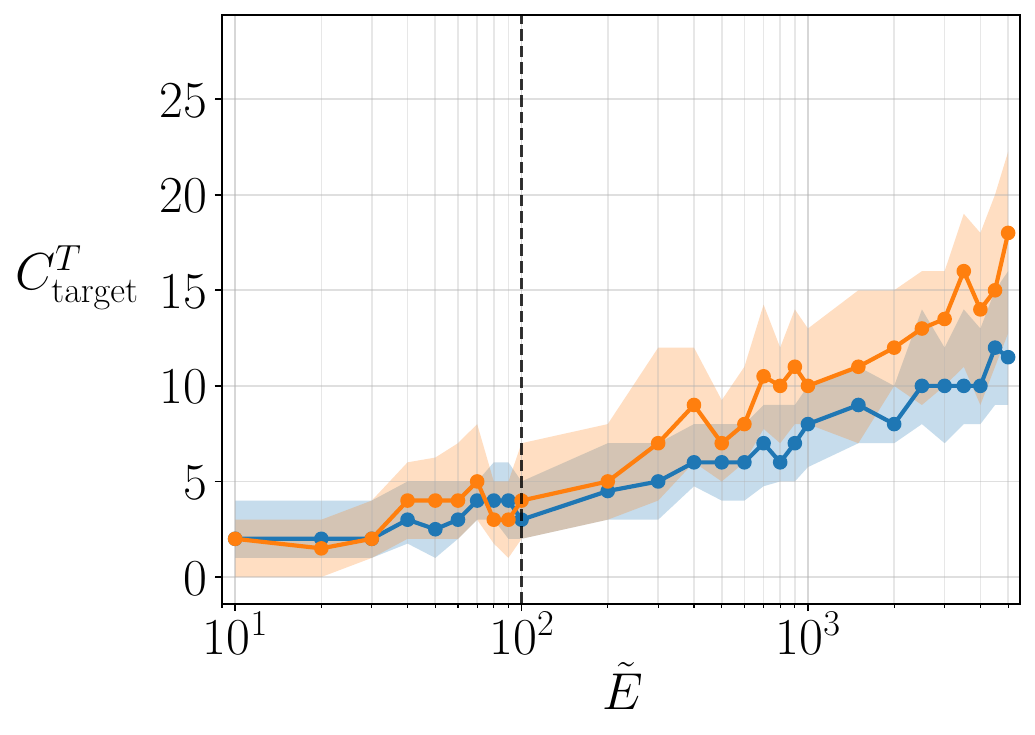}
        \caption{\(S=6\)}
    \end{subfigure}
    \hfill
        \begin{subfigure}[t]{0.24\linewidth}
        \centering
        \includegraphics[width=\linewidth]{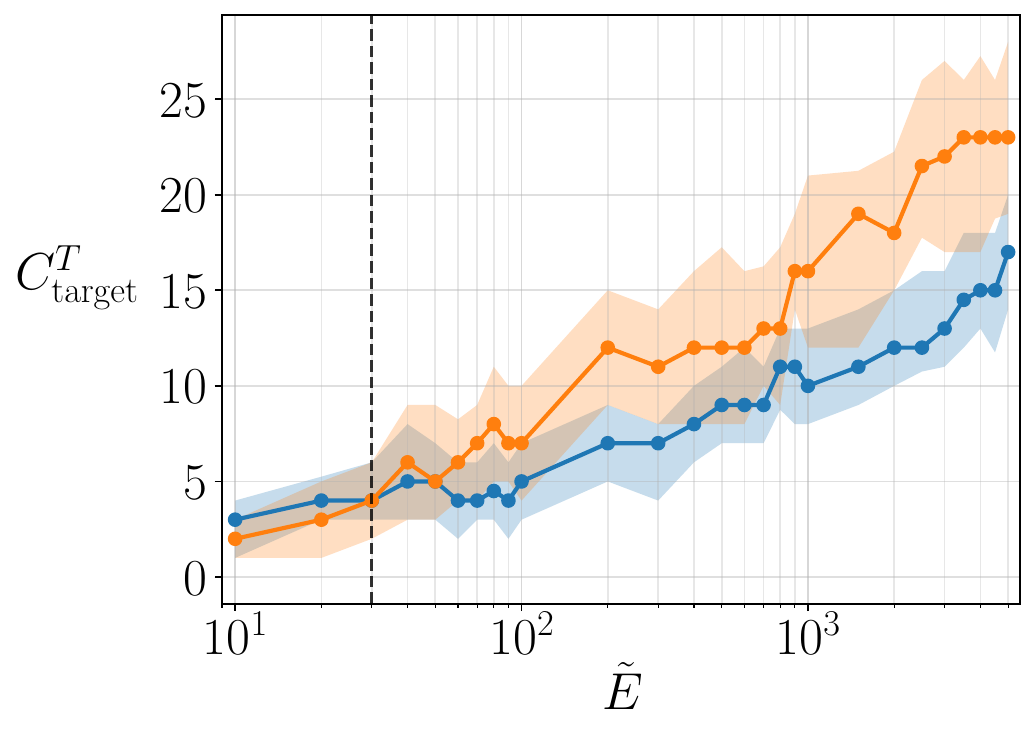}
        \caption{\(S=8\)}
        \label{fig:swarm_size_8}
    \end{subfigure}
    \caption{%
    Collected objects~\(\mathbf{C^T_\text{target}}\) for the generalist (blue) and specialist (orange) strategies for different MRS sizes~\(\mathbf{S}\) across evaluation budgets~\(\mathbf{\tilde{E}}\) over 20 independent runs per controller combination (log scale; solid line: median, shaded areas: interquartile range, black dashed line: break-even point).} 
    \label{fig:evaluation_swarm_size_normalized}
    \Description{}
\end{figure*}

Figures~\ref{subfig:generalist_10} to~\ref{subfig:collector_10} show the best fitness over generations for the generalists, the droppers, and the collectors for \(S=4\)~robots as representatives for all fitness curves. 
The main difference between the curves for different~\(S\) is in best fitness, which increases with~\(S^*\).
For the generalist behavior, we find a mean best fitness~\(F_G\) (Eq.~\ref{eq:F_G}) of \(8.4\), \(14.0\), \(20.0\), and \(26.0\) for MRS sizes~\(S\) of 2, 4, 6, and 8, respectively, in the last generation. 
The fitness curve of the generalists converges at around \(1\,000\)~generations for MRS sizes~\(S \in \{2, 4, 6\}\).
However, for \(S=8\) the fitness does not converge within the \(5\,000\) generations.
By contrast to the smaller MRS sizes, the best evolved individuals either result in the expected generalist behavior or in an emergent form of task partitioning similar to the specialist strategy (see Sec.~\ref{subsec:behaviors}).
Consequently, we find greater variance and no convergence in the fitness for \(S=8\).
For the dropper, mean best fitness~\(F_D\) (Eq.~\ref{eq:F_D}) increases from \(23.4\) (\(S^*=1\)), \(35.6\) (\(S^*=2\)) and \(44.8\) (\(S^*=3\)) to \(52.8\) (\(S^*=4\)). 
For the collector, mean best fitness~\(F_C\) (Eq.~\ref{eq:F_G}) in the last generation is \(29.2\) for \(S^*=1\), \(50.2\) for \(S^*=2\), \(63.2\) for \(S^*=3\), and \(78.0\) for \(S^*=4\).
The specialist behaviors require approximately \(2\,000\)~generations to converge.

\subsection{Post-Evaluation}

\subsubsection{Behaviors}
\label{subsec:behaviors}

We observe that the robots learn to navigate in the environment regardless of their initial positions.\footnote{Plots of the \(y\)-axis trajectories for generalist and specialist strategies for MRS size \(S=4\) are provided in the Supplementary Material (see Sec.~A).}
Generalists move back and forth between source and target after an initial transient phase.
Generalist robots learn environment-dependent motion rules (e.g., moving uphill on the slope on one side of the arena and downhill on the other).
Depending on their starting position, robots may initially even move away from the source without transporting an object. 
This behavior is explainable by the robots being memoryless and the potential loss of physical contact with an object while on the slope.
As a result, robots cannot determine whether they have already pushed an object while moving on the slope.
Three generalist controllers evolved for \(S=8\) even show a form of self-organized task partitioning similar to the specialist strategy.
Each robot only operates in a specific area of the environment and avoids crossing the slope.
Robots initialized near the source consistently behave as droppers, while robots initialized near the cache or the target behave as collectors.

In the specialist strategy, droppers and collectors first move to their respective work areas. 
Collectors reach their work area faster than droppers because robots descend the slope more quickly. 
Consequently, collectors become fully operational earlier than droppers.

\subsubsection{Varying MRS Size}
\label{subsec:swarm_size}

Next, we analyze the effect of MRS size~\(S\) on the the number of collected objects~\(C^T_\text{target}\) (Eq.~\ref{equ:c_T}) for different evaluation budgets~\(\tilde{E}\).\footnote{A plot visualizing the full statistical comparison is provided in the Supplementary Material (see Sec.~B).}
Generalists match or outperform specialists up to 
\(\tilde{E}=1\,000\) for \(S=2\),
\(\tilde{E}=300\) for \(S=4\),
\(\tilde{E}=100\) for \(S=6\), and
\(\tilde{E}=30\) for \(S=8\) (see Figs.~\ref{fig:swarm_size_2}-\ref{fig:swarm_size_8}).
Consequently, the break-even evaluation budget (see Sec.~\ref{sec:evaluation_post-evaluation}) decreases as MRS size~\(S\) increases.
For \(S=2\), we find statistically significant differences in~\(C^T_\text{target}\) after the break-even point at \(\tilde{E}=2500\) (\(p < 0.01\)).
However, this significance cannot be found for all tested larger evaluation budgets~\(\tilde{E}\).
For \(S=4\), we find first statistically significant differences for \(\tilde{E}=500\) and consistent statistically significant differences for \(\tilde{E}\ge800\) (\(p < 0.01\)).
Although we find a statistically significant difference at the break-even point (\(p = 0.044\)) for \(S=6\), this is not the case for all tested higher evaluation budgets~\(\tilde{E}\).
For \(S=8\), we find statistically significant differences for~\(\tilde{E}\ge60\) (\(p < 0.01\)).
Overall, this indicates that the evaluation budget required for specialists to outperform generalists decreases as MRS size increases.

\section{Discussion and Conclusion}

Division of labor has the potential to increase efficiency in multi-robot systems. 
However, depending on the available evaluation budget during optimization, evolved task-specialists may result in lower overall performance than optimized generalist behaviors.
In this paper, we showed that the evaluation budget required to evolve task-specialist behaviors that outperform generalists decreases with increasing MRS size. 
In future work, we plan to investigate for which MRS sizes task specialization consistently emerges without explicitly rewarding it.

\begin{acks}
PL, HH, and JK acknowledge support from DFG through Germany's Excellence Strategy-EXC 2117-422037984 and Centre for the Advanced Study of Collective Behaviour (CASCB), University of Konstanz, Konstanz, Germany.
JK acknowledges support from the Zukunftskolleg and the Carl-Zeiss-Foundation.
The authors acknowledge support by the state of Baden-Württemberg through bwHPC.
ChatGPT was used to assist drafting and refining text.
The authors critically reviewed, edited, and validated all content.
\end{acks}

\bibliographystyle{ACM-Reference-Format}
\bibliography{sample-base}

\appendix
\onecolumn

\section{Exemplary Generalist and Specialist Strategy Robot Trajectories}
\begin{figure*}[!hbt]
    \centering
    \begin{subfigure}{0.45\linewidth}
        \includegraphics[width=\linewidth]{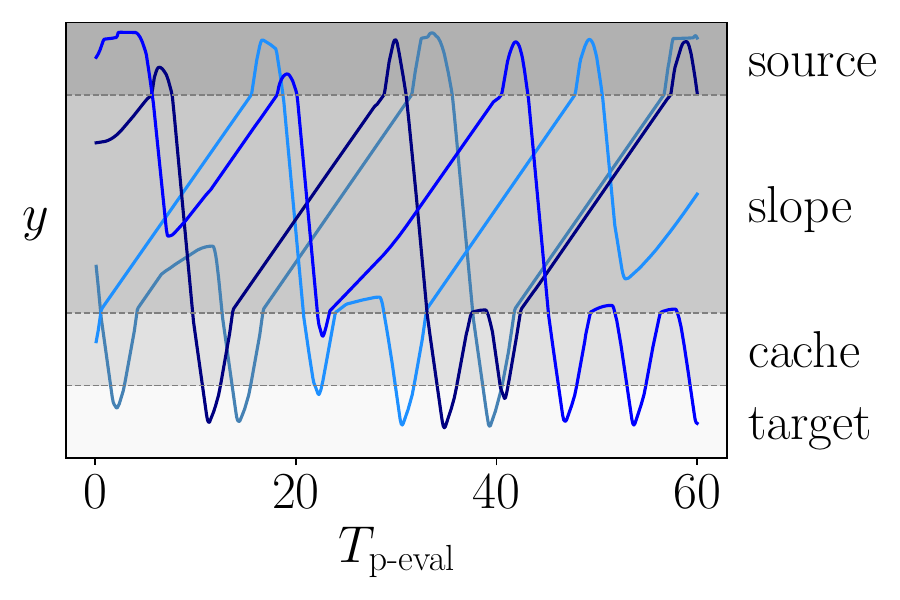}
        \caption{generalist (\(\mathbf{S=4}\))}
        \label{fig:behavior_generalist}
    \end{subfigure} 
    \hfill
    \begin{subfigure}{0.45\linewidth}
        \includegraphics[width=\linewidth]{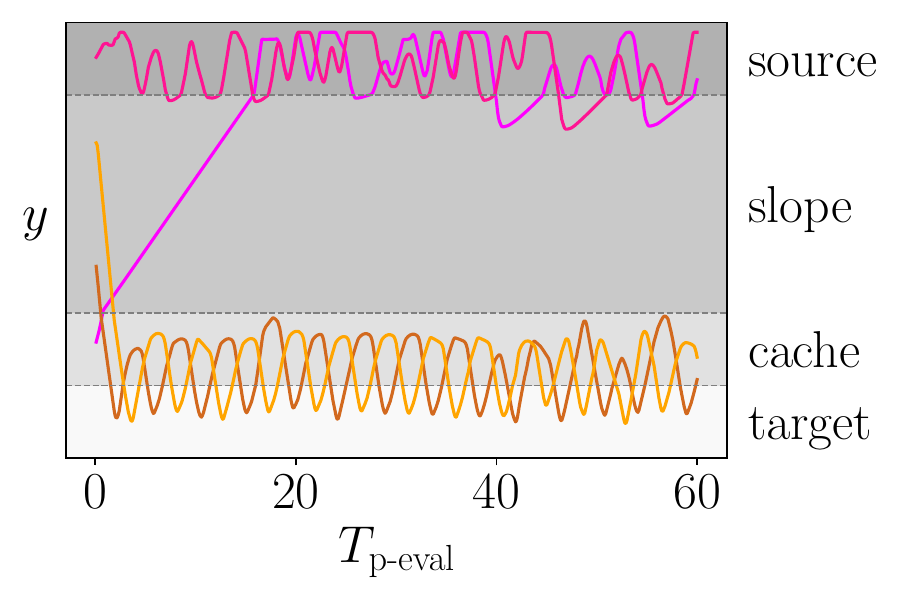}
        \caption{specialist (\(\mathbf{S=4}\))}
        \label{fig:behavior_specialist}
    \end{subfigure}
    \caption{\(\mathbf{y}\)-axis trajectories for exemplary generalist and specialist strategies over a simulation run of \(\mathbf{T_\text{p-eval}}\)~seconds with MRS size \(\mathbf{S=4}\).
    The \(\mathbf{y}\)-coordinate determines the area in which the robot is currently located (source, slope, cache, or target).
    Blue-, pink-, and orange-shaded trajectories indicate generalist, dropper, and collector behaviors, respectively. Supplementary material for Sec.~\ref{subsec:behaviors}.}
    \label{fig:robot_trajectories}
    \Description{}
\end{figure*}

\section{Statistical Comparison Between Generalist and Specialist Strategies}
\begin{figure*}[!hbt]
    \centering
    \includegraphics[width=0.75\linewidth]{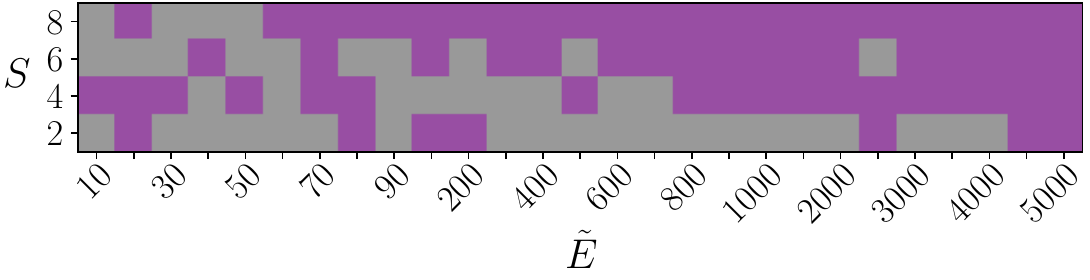}
    \caption{Statistical comparison of collected objects~\(\mathbf{C^T_{\text{target}}}\) between generalists and specialists across MRS sizes~\(\mathbf{S}\) and evaluation budgets~\(\mathbf{\tilde{E}}\). Gray: non-significance (\(\mathbf{p \geq 0.05}\)); violet: significance (\(\mathbf{p < 0.05}\)). Supplementary material for Sec.~\ref{subsec:swarm_size}.}
    \label{fig:stats_test_varying_swarm_size}
    \Description{} 
\end{figure*}

\end{document}